\documentclass[letterpaper]{article} 
\usepackage{aaai22}  
\usepackage{times}  
\usepackage{helvet}  
\usepackage{courier}  
\usepackage[hyphens]{url}  
\usepackage{graphicx} 
\usepackage{natbib}  
\usepackage{caption} 
\DeclareCaptionStyle{ruled}{labelfont=normalfont,labelsep=colon,strut=off} 
\usepackage[linesnumbered,noend]{algorithm2e}

\usepackage{subcaption}

\usepackage{tikz}
\usetikzlibrary{trees}

\usepackage{color}
\usepackage[normalem]{ulem} 

\newcommand{\bcode}{
	\begin{tabbing}
	nnn\=nnn\=nnn\=nnn\=nnn\=nnn\=nnn\=nnn\=nnn\=nnn\kill
}
\newcommand{\bncode}{
	\begin{tabbing}
	0. nn\=nn\=nn\=nn\=nn\=nn\=nn\=nn\=nn\=nn\kill
}
\newcommand{\bnncode}{
	\begin{tabbing}
	00. nn\=nn\=nn\=nn\=nn\=nn\=nn\=nn\=nn\=nn\kill
}
\newcommand{\ecode}{\end{tabbing}}

\usepackage{amsmath,amsfonts}
\usepackage{amsthm}

\newtheorem{theorem}{Theorem}
\newtheorem{lemma}{Lemma}

\newcommand{\citepw}[1]{\citeauthor{#1}~(\citeyear{#1})}

\title{Beam Search: Faster and Monotonic}

\author{Sofia Lemons,\textsuperscript{\rm 1, 2}
        Carlos Linares L\'opez,\textsuperscript{\rm 3}
        Robert C. Holte,\textsuperscript{\rm 4}
        Wheeler Ruml\textsuperscript{\rm 1}}
\affiliations {
    \textsuperscript{\rm 1} University of New Hampshire\\
    \textsuperscript{\rm 2} Earlham College\\
    \textsuperscript{\rm 3} Computer Science and Engineering Department, Universidad Carlos III de Madrid\\
    \textsuperscript{\rm 4} University of Alberta, Alberta Machine Intelligence Institute (Amii)\\
    sofia.lemons@earlham.edu, carlos.linares@uc3m.es, rholte@ualberta.ca, ruml@cs.unh.edu
}

\begin{document}

\maketitle

\begin{abstract}

Beam search is a popular satisficing approach to heuristic search problems that allows one to trade increased computation time for lower solution cost by increasing the beam width parameter.  We make two contributions to the study of beam search.  First, we show how to make beam search monotonic; that is, we provide a new variant that guarantees nonincreasing solution cost as the beam width is increased.  This makes setting the beam parameter much easier.  Second, we show how using distance-to-go estimates can allow beam search to find better solutions more quickly in domains with non-uniform costs.  Together, these results improve the practical effectiveness of beam search.

\end{abstract}

\section{Introduction}

Heuristic state-space search is a fundamental problem-solving methodology.  The search frontier of open nodes in a best-first search such as A* or weighted A* can, and typically does, grow very large.  This limits the scalability of best-first search for solving large or difficult problems.  Beam search \cite{bisianiEAI1987} can be seen as one way to adapt best-first search to large problems.

Beam search takes as parameters the start state and beam width and searches the state space level by level, as in breadth-first search. If more nodes are generated for the next level than allowed by the beam width, only those with the lowest f-values are retained.  (We assume throughout this paper that the heuristics being used are admissible, but not necessarily consistent.) Ties are broken in preference of nodes with lower h-values.  The algorithm stops after completing the level at which a solution is found, returning the lowest cost solution from that level if there are multiple. 

\begin{figure}[t]
	\begin{center}
	\includegraphics[width=3.25in]{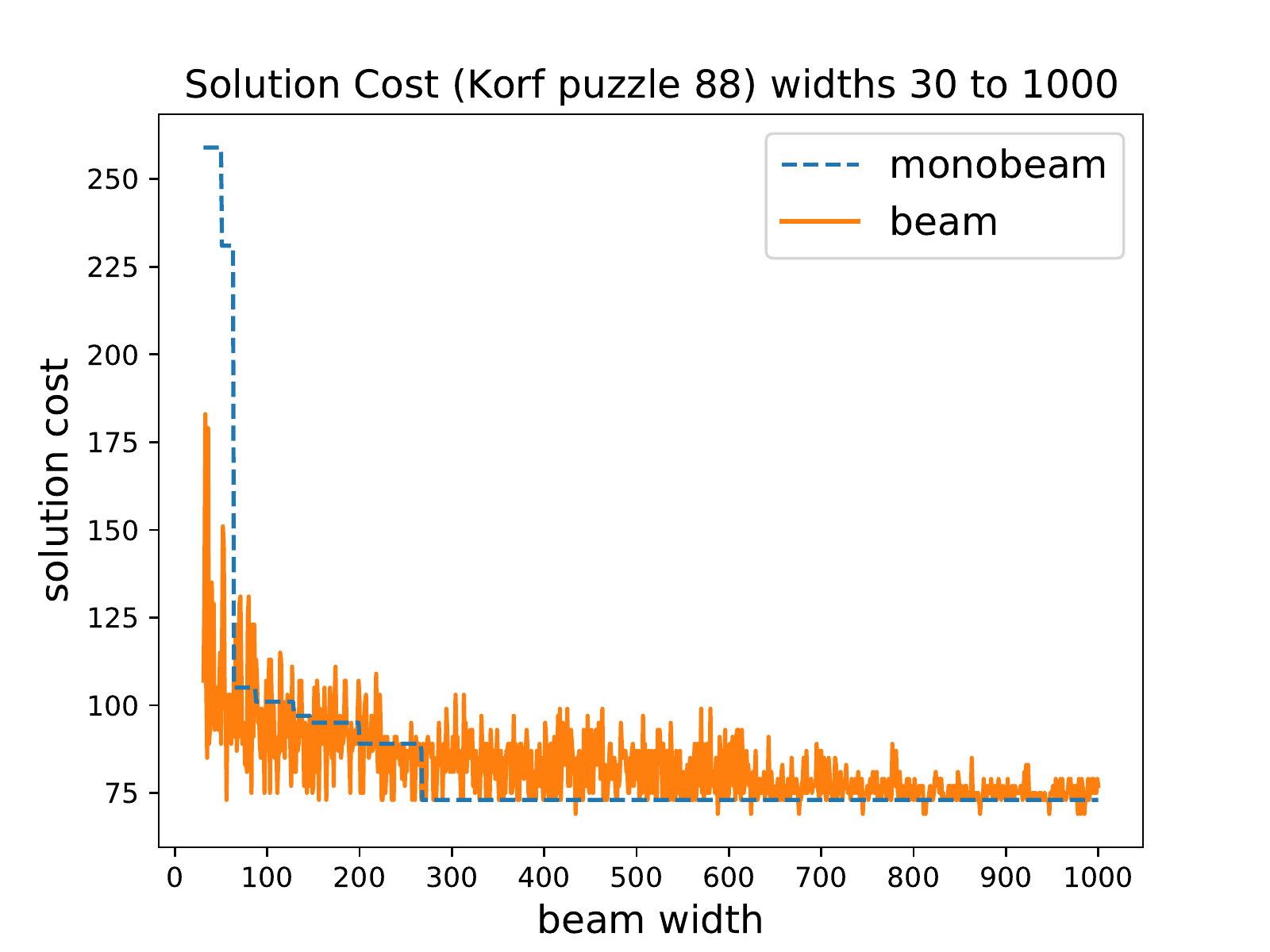}
	\end{center}
	\vspace{-.15in}
	\caption{Solution cost as beam width varies.}
	\label{fig:non-monotonic-plot}
\end{figure}

Despite its apparent simplicity, the behavior of beam search is not well-understood and can be surprising.  For example, increasing the beam width does not always result in finding better solutions. Figure \ref{fig:non-monotonic-plot} demonstrates this non-monotonic behavior of beam search on an instance of the 15-puzzle.  As the beam width is increased, the general trend is for solution cost to decrease, but this does not necessarily hold for specific beam widths.  To gauge this phenomenon quantitatively, we will call a beam width $k>1$ ill-behaved if the cost of the solution that beam search returns when run with width $k$ is higher than the cost of the solution found when using $k-1$.  When looking at all widths from 30 through 1000 on \citepw{korf:ida}'s 100 15-puzzle instances, a full 30\% of the beam widths are ill-behaved on average (varying from 5\% to 44\% depending on the instance, with a median of 31\%).  If one considers the three outcomes to be higher, lower, or staying the same, this is almost random chance.  A practitioner trying to solve a problem needs to know whether to increase or decrease the specific beam width that they are using if they wish to decrease solution cost, so this behavior of beam search can be very annoying.  In this paper, we introduce an algorithm, {\bf monobeam}, that provably eliminates this annoying behavior.  Its performance is also shown in Figure~\ref{fig:non-monotonic-plot}.  We show how this can be done even in combination with duplicate elimination.

In Figure~\ref{fig:non-monotonic-plot}, we also see that there is not necessarily a large price paid, in terms of solution cost, to obtain the guarantee of monotonicity. Indeed, there are many beam widths in this plot for which monobeam's solution is cheaper than beam's.  The experiments in this paper are largely aimed at investigating this question of what price is paid for monotonic behavior. We demonstrate an approach that can select nodes for the beam in a way that preserves monotonicity, without incurring additional overhead in the selection process.

Furthermore, in addition to its nonmonotonicity, beam search demonstrates undesirable behavior in our non-unit-cost problem domains, often finding very poor solutions or no solutions at all.  We explore the use of distance-to-go estimates with both regular and monotonic beam search, finding that they ameliorate the poor behavior of beam search in non-unit-cost domains.  We discover, in multiple domains, the surprising result that using the distance-to-go measurement not only improves search time but often reduces solution cost as well. 

By providing beam search variants that are monotonic and effective with non-unit costs, this work makes beam search easier to apply in practice.

\section{Previous Work}

Beam search is a popular heuristic search method and has been widely used in several fields, including natural language processing \cite{cohen:eab,meister:cps} and operations research \cite{valente:bsa}.  Wilt, Thayer, and Ruml (\citeyear{wilt:cgs,wilt:sgs}) provide empirical comparisons of various flavors of beam search, finding that the breadth-first-based variant that we use here is preferred.

The most popular research direction regarding beam search has probably been completeness.  \citepw{zhang:cab} proposes a complete anytime beam search based on iteratively weakening pruning rules that exclude nodes from the beam.
\citepw{zhou:bss} make beam search complete by integrating systematic backtracking. 
\citepw{furcy:ldb} use limited discrepancy search instead of chronological backtracking.  Our concerns in the current paper are orthogonal to completeness: we aim to make beam search itself more effective and any of those completeness methods could be layered on top.

Some research has investigated finding a suitable beam width, observing that larger beam widths might incur larger computational costs while they are more likely to avoid errors in node selection~\cite{ow:extended,wijs:ebs}. Monotonicity makes it easier to select the right beam width, because a user can pick a beam width as large as their system is able to accommodate, without worrying about whether that width will cause it to perform worse than it would have at lower widths.

Nonmonotonicity of beam search has been recognized before.  For example, \citepw{cohen:eab} (see also citations therein) aim to explain the degradation in the performance of beam search with larger beam widths, but from the perspective of heuristic search.  \citepw{vadlamudi:ibs} propose an algorithm, Incremental Beam Search, which when given a maximum width and a maximum depth, iteratively performs beams searches of increasing width.  The notable feature of their method is that it avoids repeated work by storing the previous beam search so that the next increment of width can be performed quickly.  The algorithm is anytime, in that solutions found by earlier widths are available as the search widens to the pre-specified maximum width.  However, it stops at that maximum width and does not converge to optimal solutions.  In addition to the beam width, the maximum search depth must be specified for the method in advance, making it difficult to use in practice as reasonable bounds are often unknown before a problem is solved. Furthermore, the algorithm is only monotonic across the beam widths within a single execution, not when rerun with a larger maximum beam width. Preliminary experiments on unit 15-puzzle instances showed that providing anything other than a rather tight depth bound led to uncompetitive runtimes. For these reasons, we do not consider this algorithm further.

\section{Monotonic Beam Search}

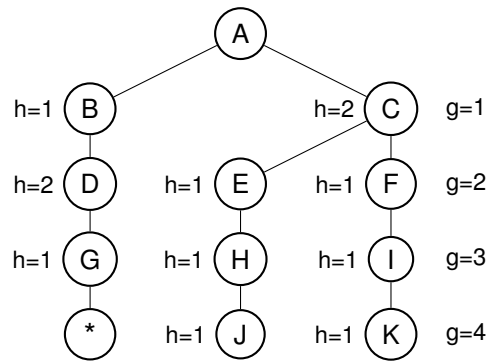
\begin{figure}[t!]
\tikzset{
    every node/.style={font=\sffamily\small},
    main node/.style={thick,circle,draw,font=\sffamily\normalsize}
}
\begin{tikzpicture}[-, sibling distance=2cm, level distance=1cm]
    \node[main node]{A}
    child {
        node[main node][label=left:{h=1}]{B}
        child {
            node[main node][label=left:{h=2}]{D}
            child {
                node[main node][label=left:{h=1}]{G}
                child {
                    node[main node, font=\sffamily\Large]{*}
                }
            }
        }
    }
    child[missing]{}
    child {
        node[main node][label=left:{h=2}]{C}
        child {
            node[main node][label=left:{h=1}]{E}
            child {
                node[main node][label=left:{h=1}]{H}
                child {
                    node[main node][label=left:{h=1}]{J}
                }
            }
        }
        child {
            node[main node][label=left:{h=1}]{F}
            child {
                node[main node][label=left:{h=1}]{I}
                child {
                    node[main node][label=left:{h=1}]{K}
                }
            }
        }
        child [grow=right] {
        node {g=1} edge from parent[draw=none]
        child [grow=down] {
        node {g=2} edge from parent[draw=none]
        child [grow=down] {
        node {g=3} edge from parent[draw=none]
        child [grow=down] {
        node {g=4} edge from parent[draw=none]
        }
        }
        }
        }
    };
\end{tikzpicture}
\caption{A search space in which non-monotonicity occurs when the beam width is increased from~1 to~2.}
\label{fig:non-monotonic_example}
\end{figure}

One possible cause of beam search's non-monotonic behavior is that, when increasing the width of the beam, there is a risk that the nodes expanded at one of the later positions in the beam may dominate the priority queue used to select nodes for the next beam and displace nodes that would have been selected with a smaller beam width. The children of nodes generated at the larger beam width may have incorrectly low cost-to-go estimates due to heuristic error. We refer to these as {\em cuckoo nodes} because they cause truly better nodes to be pushed off the beam, just as, when a cuckoo bird lays its eggs in another bird's nest, the cuckoo chicks push the other bird's eggs out of the nest. The displaced nodes may look less promising now but may be the ones which led the narrower beam width search to a better solution than will be found via the cuckoo nodes. In general, this can cause beam search, with a larger beam width, to do more work to find a goal, 
to find a solution of poorer quality or even to fail to find a solution when a solution is found with a smaller beam width.

An example is given in Figure~\ref{fig:non-monotonic_example}, where a beam search with width~2 returns a solution of lower quality than a beam search with width~1. The node indicated by * is a solution.
With width~1, beam search chooses node~B at level~1 and proceeds via node~D to find the solution beneath node G. With width~2, nodes B and C are chosen at level~1, but at level~2 node~D is displaced by C's two children and it is no longer possible to reach a solution as cheap as that found beneath node~D.

\subsubsection{Monotonicity}

\begin{algorithm}[t!]
\Begin{
solutionCost $\leftarrow \infty$\;
beam[1] $\leftarrow$ start\;
\While {at least one slot in the beam has a node with f-value $<$ solutionCost \label{lin:stop}}{
candidates $\leftarrow \emptyset$, nextBeam $\leftarrow [ ] $\;
\For{each beam slot c from 1 to width \label{lin:cloop}}{
\If {beam[c] is a node}{
\For{each child of beam[c]}{ \label{lin:expand}
\If{f(child) $<$ f(beam[c]) \label{lin:pathmax}}{
f(child) $\leftarrow$ f(beam[c])\;
} \label{lin:setpathmax}
\If {child is a goal and f(child) $<$ solutionCost \label{lin:goaltest}}{
store as solution\;
solutionCost $\leftarrow$ f(child)\;
}
\Else{
add child to candidates\; \label{lin:addtocandidates}
}
}
}
\If {candidates is nonempty \label{lin:candidatesnotempty}}{
nextBeam[c] $\leftarrow$ remove $\min f$-value node from candidates\; \label{lin:addtobeam} \label{lin:removefromcandidates}
}
}
beam $\leftarrow$ nextBeam\; \label{lin:setbeam}
}
return solution\;
}
\caption{monobeam(start,width)}\label{alg:monobeam}
\end{algorithm}

In order to avoid cuckoo nodes as the beam is widened, we consider each specific position in the beam, or {\em slot}, sequentially.  Pseudocode for our method, {\bf monobeam}, is shown in Algorithm~\ref{alg:monobeam}.  It maintains a priority queue, \emph{candidates}, containing (line~\ref{lin:addtocandidates}) only children of nodes at the current beam slot and slots before it. For each slot on the beam, it expands the current node in that slot (line~\ref{lin:expand}) and adds to \emph{candidates} the non-goal children (lines~\ref{lin:goaltest}--\ref{lin:addtocandidates}). It then pops the minimum $f$-valued node from \emph{candidates} and stores it in the current slot of nextBeam (line~\ref{lin:addtobeam}). Ties among minimum $f$-valued nodes are broken on low $h$-values. This means that the selection of the node for a slot is not affected by any later slots.

A second possible cause of beam's non-monotonic behavior is its stopping condition. It stops at the shallowest level where it finds a solution, even if that beam contains nodes with $f$-values smaller than the solution's cost.  A smaller beam width might have found a better solution deeper in the search beneath one of the nodes with the smaller $f$-values. To preserve monotonicity, monobeam continues the search until no nodes are left on the beam with f-values lower than the current incumbent solution, recording any lower cost solutions found. To ensure we are always getting closer to this stopping condition, pathmax~\cite{mero:pathmax} is used to ensure $f$-values along a path are non-decreasing (lines~\ref{lin:pathmax}--\ref{lin:setpathmax}.) We also assume that the domain does not contain infinite sequences of zero-cost operators.

Note that Algorithm~\ref{alg:monobeam} does not detect duplicates and retains nodes on the beam whose f-values are worse than the incumbent solution. We will address those issues below, but first we will prove some properties of Algorithm~\ref{alg:monobeam}.

\begin{lemma}
\label{lemma:slotselection}
Given the beam at the start of any iteration of the while loop at line~\ref{lin:stop} of Algorithm \ref{alg:monobeam}, at the end of the $c$-loop at line~\ref{lin:cloop} the node in nextBeam[c] for all $c$ will be the same for all $\mathit{width} \geq c$.
\end{lemma}
\begin{proof}
By induction on $c$.
When $c=1$ this is trivially true because only the children of the node in slot 1 of beam have been added to candidates at line \ref{lin:addtocandidates} when a node for nextBeam[1] is selected at line~\ref{lin:addtobeam}. We will always select the minimum value child of the node in beam[1] to fill nextBeam[1] regardless of beam width and the node selected for nextBeam[1] will be removed from candidates.

For any slot $c>1$, the inductive hypothesis asserts that the same nodes were selected to fill nextBeam slots 1 through $c-1$ regardless of beam width. We will now show this implies the same node will be selected for nextBeam slot $c$ regardless of beam width. At the time this selection is made (line~\ref{lin:addtobeam}), candidates contains the non-goal children of all the beam nodes in slots 1 to $c$ except those selected for nextBeam's slots 1 to $c-1$. The minimum value node from this set will be selected to fill nextBeam slot $c$. The nodes in candidates are therefore independent of the beam width, and therefore so is the node selected for nextBeam slot $c$.
\end{proof}

\begin{lemma}\label{lemma:identicallevels}
Let $w_a < w_b$ be beam widths, $d_i$ the depth at which monobeam with width $w_i$ terminates, and $beam_{i,d}$ the beam when the stopping condition is tested at the beginning of the while loop for iteration $d$ when width $w_i$ is being used. Then for all depths $1 \dots \min\{d_a,d_b\}$, $beam_{a,d}[j] = beam_{b,d}[j]$ for all $1 \le j \le w_a$.
\end{lemma}

\begin{proof}
By induction on $d$. When $d=1$, $beam_{a,1} = beam_{b,1}$; both searches insert start in slot~1 and leave empty slots everywhere else.  For any depth $d > 1$, the inductive hypothesis asserts that $beam_{a,d-1}[j] = beam_{b,d-1}[j]$ for all $1 \le j \le w_a$. Applying Lemma~\ref{lemma:slotselection}, we get that, at the end of iteration $d-1$, $nextBeam[j]$ is independent of the beam width for all $1 \le j \le w_a$. The lemma follows because $beam[i,d]$ is $nextBeam$ at the end of iteration $d-1$.
\end{proof}

\begin{theorem}
\label{theo:monotonicity}
Algorithm \ref{alg:monobeam} with an admissible heuristic and beam width $k+1$ will always return a solution with cost lower than or equal to the cost of the solution it would have returned if it had been run with beam width $k$, for all $k \geq 1$.
\end{theorem}
\begin{proof}
There are three cases to consider.
First, a solution may be found in slot $k+1$ with cost lower than the solution found by a search with beam width $k$. This solution will be kept as an incumbent until all nodes on the beam have f-values greater than or equal to its cost (line \ref{lin:stop}), at which time it will be returned. Because it has a lower cost than the solution found by a search with beam width $k$, the theorem holds.

Second, if no solution is found in slot $k+1$, we know from Lemma \ref{lemma:identicallevels} that the nodes selected for slots 1 through $k$ will be unaffected by the increased beam width. Thus, the solution found by a width $k$ search will be found in these slots and returned.

Third, if a solution is found at beam slot $k+1$ with cost greater than the solution that would have been found by a search with width $k$, we know from Lemma \ref{lemma:identicallevels} that we will still select in slots $1$ through $k$ all nodes which would be selected by a search of width $k$. We will find the lower-cost solution which would have been found by a search with width $k$ and we will return that solution.
\end{proof}

\subsection{Pruning based on incumbent solutions}

\begin{algorithm}[t!]
\Begin{
\For{each beam slot c from 1 to width}{
\If{f(nextBeam[c]) $\geq$ solutionCost \label{lin:ftest}}{
nextBeam[c] $\leftarrow$ empty\;
}
}
}
\caption{Pruning based on incumbent solution, to be added before line \ref{lin:setbeam} of Algorithm~\ref{alg:monobeam}.}\label{alg:pruning}
\end{algorithm}

\begin{lemma}
\label{lemma:nogoodremoved}
Algorithms~\ref{alg:monobeam} and \ref{alg:pruning} with an admissible heuristic will not remove any node that is along a path to a solution with cost lower than the incumbent's.
\end{lemma}
\begin{proof}
Algorithm~\ref{alg:pruning} excludes nodes from the beam with $f$-values $\geq$ the current incumbent solution cost.  Because the heuristic is admissible, such nodes cannot lead to a solution better than the current incumbent.
\end{proof}

Pruning can reduce the number of expansions required, as well as the size of candidates throughout the search. It might mean that we do not explore exactly the same set of nodes when the beam width expands, because we might find an incumbent solution at a larger beam width and prune away nodes from lower beam slots. However, monotonicity can still be maintained.

\begin{lemma}
\label{lemma:pruningsame}
Algorithms \ref{alg:monobeam} and \ref{alg:pruning} with an admissible heuristic will select for nextBeam at each level exactly those nodes that would have been selected by Algorithm 1 alone, except possibly omitting some nodes which have f-values $\geq$ the incumbent's.
\end{lemma}
\begin{proof}
First, note that pruning does not directly introduce any new nodes into the beam.  Second, consider the consequences of pruning a node $n$ (because its $f$-value is not better than the incumbent solution's). We know by Lemma \ref{lemma:nogoodremoved} that it could not have led to a better solution than the current incumbent. Its children must (by lines \ref{lin:pathmax} and \ref{lin:setpathmax} of Algorithm \ref{alg:monobeam}) have $f$-values no better than the incumbent solution.  If those children would have been selected to fill any given slot in nextBeam without pruning, then candidates at that time did not contain any nodes with $f$-values better than the incumbent, so that slot will remain empty when pruning.  So pruning does not introduce nodes or cause nodes to be selected for different slots than they would have been otherwise. We will select for each slot on nextBeam exactly those nodes selected by Algorithm \ref{alg:monobeam} that have $f$-values less than the incumbent's.
\end{proof}

\begin{theorem}
\label{theo:pruning}
Algorithms \ref{alg:monobeam} and \ref{alg:pruning} with an admissible heuristic and beam width $k+1$ will always return a solution with cost $\leq$ to the cost of the best solution found by the same algorithm with beam width $k$, for all $k \geq 1$.
\end{theorem}
\begin{proof}
There are three cases.  First, if no solution is found at slot $k+1$, then all search in slots $1$ through $k$ will proceed identically to a search of width $k$, and the same solution will be returned as would be found by a search of width $k$.
Second, if a solution is found at slot $k+1$ with cost less than the solution that would be found by a search with width $k$, we will eventually explore all nodes with f-values less than the incumbent solution's cost, and will return the solution found at width $k+1$.
Third, if a solution is found at slot $k+1$ with cost greater than the solution which would be found by a search of width $k$, we know from Lemma \ref{lemma:pruningsame} that we can use that incumbent solution to prune nodes with f-values $\geq$ the cost of that incumbent while still maintaining all the nodes that could lead to a better solution. Thus, we will eventually find the solution that would have been found by a search of width $k$ and return it.
\end{proof}

\subsection{Duplicate Handling}

\begin{algorithm}[t!]
\Begin{
done $\leftarrow$ false\;
\While{candidates is nonempty and \\ done = false}{
node $\leftarrow \min f$-value node from candidates\;
node.width $\leftarrow$ c\;
\If{node's state is not in closed list}{
add node to closed\;
done $\leftarrow$ true\;
}
\Else{
dup $\leftarrow$ closed entry for node's state \;
\If{node.width $<$ dup.width \label{lin:widthtest}}{
update closed entry to node\;
done $\leftarrow$ true\;
}
\ElseIf{f(node) $\leq$ f(dup)}{
\If{node.width = dup.width}{
update closed entry to node\;
}
done $\leftarrow$ true\;
}
\Else{
remove node from candidates\;
}
}
}
}
\caption{Duplicate elimination, to be added before line \ref{lin:candidatesnotempty} of Algorithm \ref{alg:monobeam}}\label{alg:dedup}
\end{algorithm}

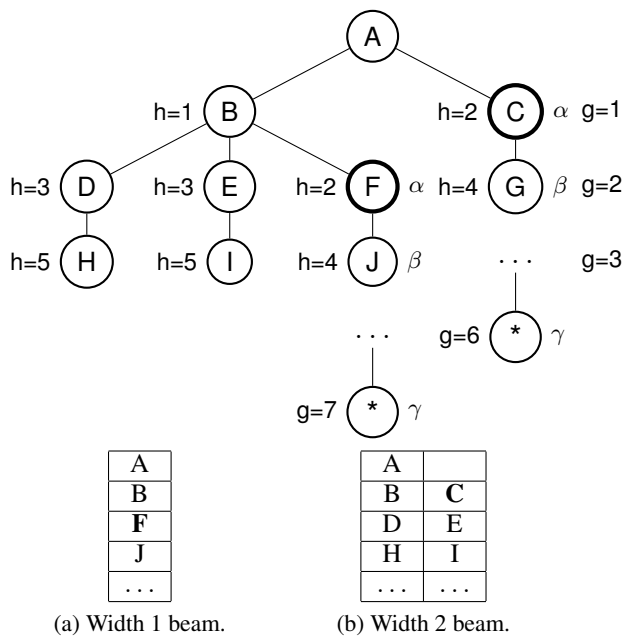
\begin{figure}[t!]
\tikzset{
    every node/.style={font=\sffamily\small},
    main node/.style={thick,circle,draw,font=\sffamily\normalsize},
    bold node/.style={ultra thick,circle,draw,font=\sffamily\normalsize}
}
\begin{subfigure}{3in}
\begin{center}
\begin{tikzpicture}[-, sibling distance=1.9cm, level distance=1cm]
    \node[main node]{A}
    child {
        node[main node][label=left:{h=1}]{B}
        child {
            node[main node][label=left:{h=3}]{D}
            child {
                node[main node][label=left:{h=5}]{H}
            }
        }
        child {
            node[main node][label=left:{h=3}]{E}
            child {
                node[main node][label=left:{h=5}]{I}
            }
        }
        child {
            node[bold node][label=left:{h=2}, label=right:$\alpha$]{F}
            child {
                node[main node][label=left:{h=4}, label=right:$\beta$]{J}
                child {
                    node[font=\sffamily\Large]{$\hdots$} edge from parent[draw=none]
                    child {
                        node[main node, label=left:{g=7}, label=right:$\gamma$][font=\sffamily\Large]{*}
                    }
                }
            }
        }
    }
    child[missing]{}
    child {
        node[bold node][label=left:{h=2}, label=right:$\alpha$]{C}
        child[missing]{}
        child {
            node[main node][label=left:{h=4}, label=right:$\beta$]{G}
                child {
                    node[font=\sffamily\Large]{$\hdots$} edge from parent[draw=none]
                    child {
                        node[main node, label=left:{g=6}, label=right:$\gamma$][font=\sffamily\Large]{*}
                    }
                }
        }
        child [grow=right, level distance=1.15cm] {
        node {g=1} edge from parent[draw=none]
        child [grow=down, level distance=1cm] {
        node {g=2} edge from parent[draw=none]
        child [grow=down] {
        node {g=3} edge from parent[draw=none]
        }
        }
        }
    };
\end{tikzpicture}
\end{center}
\end{subfigure}

\vspace{0.05in}

\begin{subfigure}{1.45in}
\begin{center}
\begin{tabular}{ | c | }
\hline
A \\
\hline
B \\
\hline
{\bf F} \\
\hline
J \\
\hline
$\hdots$ \\
\hline
\end{tabular}
\end{center}
\vspace{-0.1in}
\caption{Width 1 beam.}
\end{subfigure}
\begin{subfigure}{1.45in}
\begin{center}
\begin{tabular}{ | c | c | }
\hline
A & \\
\hline
B & {\bf C} \\
\hline
D & E \\
\hline
H & I \\
\hline
$\hdots$ & $\hdots$  \\
\hline
\end{tabular}
\end{center}
\vspace{-0.1in}
\caption{Width 2 beam.}
\end{subfigure}
\caption{An example in which full-beam duplicate elimination causes monobeam to behave  non-monotonically.}
\label{fig:dedup-non-monotonic-example}
\end{figure}

Eliminating duplicate nodes during search can be important for search spaces with many cycles and transpositions, but must be done with care to preserve monobeam's monotonicity.
Like regular beam search, for the purpose of duplicate elimination, monobeam stores in a closed list all the nodes that have been part of any beam along with their $f$-values.
Beam search's method of duplicate elimination, which we call full-beam duplicate elimination, is: if a node is generated with $f(n)=f_{new}$ and it occurs in closed with $f(n)=f_{old}$ then it is discarded if $f_{new} \ge f_{old}$.

To see how adding this to monobeam can introduce non-monotonicity, consider the example
in Figure \ref{fig:dedup-non-monotonic-example} where a monobeam search using full-beam duplicate elimination with width~2 will return no solution, whereas a monobeam search with width~1 will return a solution. In this figure, nodes~C and~F represent the same state, which we will call $\alpha$, reached by different sequences of actions.  The trees below them are, of course, identical (for example, nodes~G and~J represent the same state, $\beta$, and the goals below them are the same state $\gamma$).  There are no solutions below nodes~H and~I. The beams at each level for widths~1 and~2 are shown in the boxes at the bottom of the figure.

For beam width~1, monobeam proceeds directly to the goal via path $A - B - F - J \dots \gamma$. The width~2 search puts a node representing $\alpha$ into its level~1 beam when it selects node~C.  This causes $\alpha$ to be put into the closed list with $f(\alpha)=3$.
When the node in slot~1 of this beam (B) is expanded, D, E, and F are generated. Because the state F represents is already closed with a smaller $f$-value, F is dropped. One of D or E is placed in slot~1 of the next beam and the other remains as a candidate for slot~2. Because $f(G)$ is larger than that candidate's $f$-value, the beam for the next level consists of $D$ and $E$ (in some order), and not $G$. There is now no path to a solution.

For this reason, duplicate handling must also pay attention to the slot from which a node was expanded in order to preserve monotonic behavior. Nodes that were expanded from higher beam slots cannot be counted as duplicates in the lower beam slots or else we risk pruning away a solution that would have been reached at that lower width. Therefore, we propose a method of duplicate elimination in which the closed list records the beam slot at which a node was expanded (denoted as $width$ in Algorithm \ref{alg:dedup}) and that node can only be considered as a duplicate at that width and higher (line~\ref{lin:widthtest}). If we see the node again at a lower beam slot, we update the entry in the closed list to change it to the lower width value. Even if the higher beam slot version has a better cost, we must use the worse cost version for pruning in lower beam slots in order to preserve the search behavior that they would have followed if the higher beam slots had not existed. It is possible to maintain both versions for multiple widths to improve pruning, but this would increase both the computational and memory requirements of duplicate checking.

\begin{theorem}
\label{duplicates}
Algorithms \ref{alg:monobeam} and \ref{alg:dedup} with an admissible heuristic and beam width $k+1$ will always return a solution with cost lower or equal to the cost of the best solution found by the same algorithm with beam width $k$, for all $k \geq 1$.
\end{theorem}
\begin{proof}
Because line \ref{lin:widthtest} in Algorithm \ref{alg:dedup} provides that a duplicate child of a node from slot $k$ can only be eliminated if the original copy was expanded from one of the slots 1 through $k$, then Lemma \ref{lemma:identicallevels} still holds and the proof is identical to the proof of Theorem \ref{theo:monotonicity}.
\end{proof}

\subsubsection{A Disadvantage of Monotonicity}

Monotonicity limits the pool of nodes from which the search can choose as it selects nodes for each slot of the next layer's beam.  In its attempt to avoid cuckoo nodes, monobeam prevents the search from choosing exactly those children of the entire current beam that have the lowest $f$-values (regardless of parentage).  If the heuristic is generally helpful, then these constraints will tend to lead to solutions with increased cost compared to beam.  It is an empirical question how severe the price of monotonicity tends to be.

\subsection{Empirical Analysis}

We implemented beam and monobeam in C++ \footnote[1]{Code available at \url{https://github.com/snlemons/search}.} and tested their behavior on several classic search benchmarks. The implementation of monobeam tested uses the pruning technique given in Algorithm \ref{alg:pruning} and the duplicate elimination technique given in Algorithm \ref{alg:dedup}. Each algorithm was run with widths 30, 100, 300, 1000, 3000, 10000, 30000, and 100000. Algorithms were given a memory limit of 7.5GB.

\subsubsection{Sliding Tile Puzzle}

We used five cost models: unit cost, where the cost of moving any tile is 1; heavy cost, moving tile numbered $t$ costs $t$; sqrt cost, moving tile $t$ costs $\sqrt{t}$; inverse cost, $1/t$; reverse cost, moving tile t costs $16 - t$.  The cost-to-go heuristic was a weighted version of the Manhattan distance in which each tile's Manhattan distance is multiplied by the cost of moving that tile.  Our implementation expands nodes at a rate of approximately 1.5 million nodes per second.  The standard \citepw{korf:ida} 100 15-puzzles were used in all cost models.

First, we verify that monobeam's behavior is in fact monotonic, as claimed by Theorem~\ref{theo:monotonicity}.  Figure~\ref{fig:non-monotonic-plot} illustrates its behavior, showing that monobeam lives up to its name.  It often finds worse solutions that regular beam search for very low beam widths, but its steady behavior can be advantageous as the beam width increases.  Whereas we saw in the introduction that beam search was ill-behaved on average 30\% of the time, of course monobeam is never ill-behaved.
In this sense, monobeam can make beam-style searches simpler to use, because we can pick a beam width as large as our system is able to accommodate, without worrying about whether that width will cause it to perform worse than it would have at lower widths.

\def \plotwidth {2.6in}
\def \plotadjust {-0.3in}
\def \captionadjust {-0.2in}

\newcommand{\incfig}[1]{
\includegraphics[width=\plotwidth{}]{#1}
}

\begin{figure*}[tp!]
\begin{subfigure}{\plotwidth{}}
\incfig{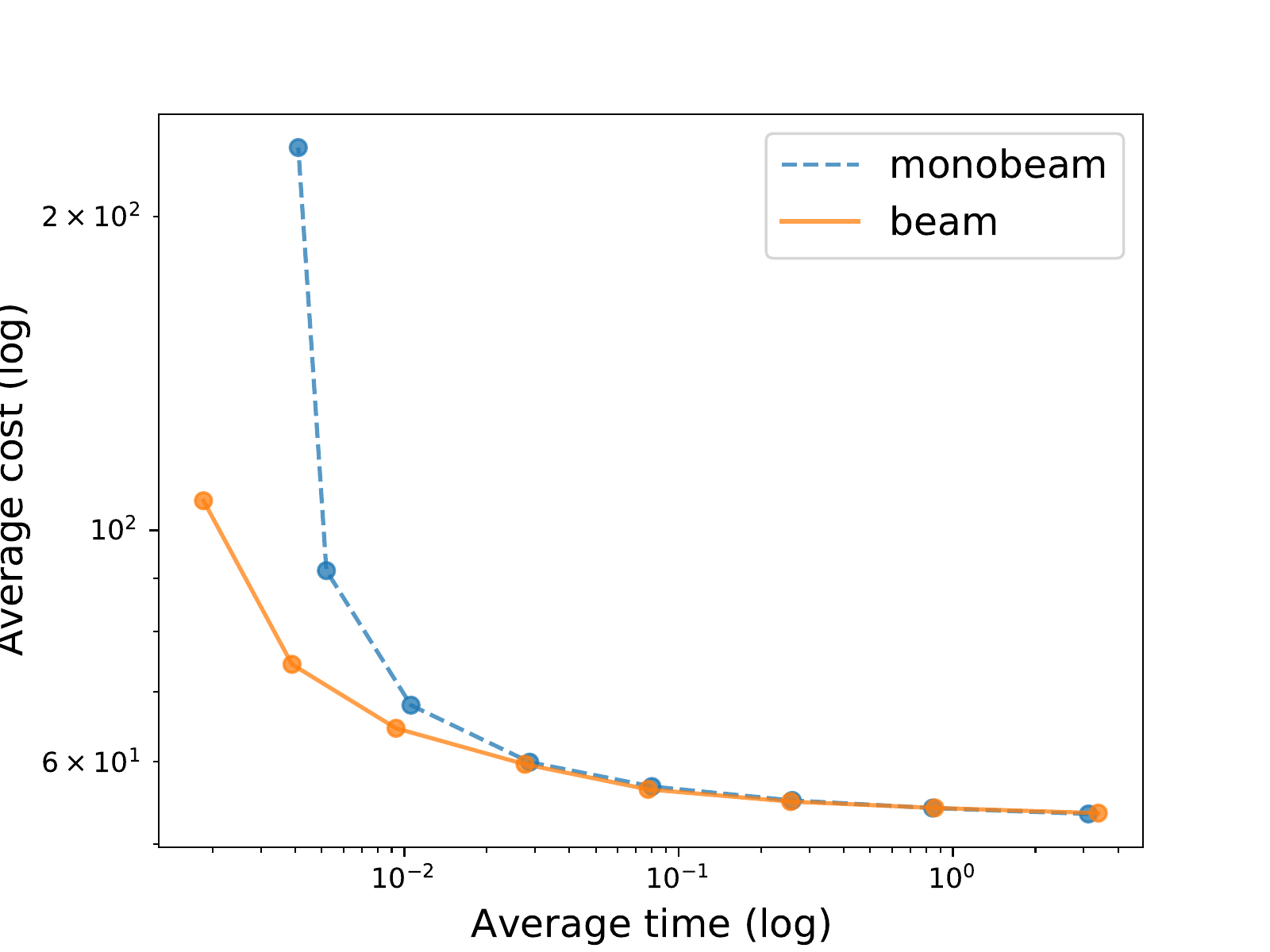}
\vspace{\captionadjust}
\caption{\small{15-puzzle (unit cost)}}
\label{fig:tile-average-unit}
\end{subfigure}
\hspace{\plotadjust{}}
\begin{subfigure}{\plotwidth{}}
\incfig{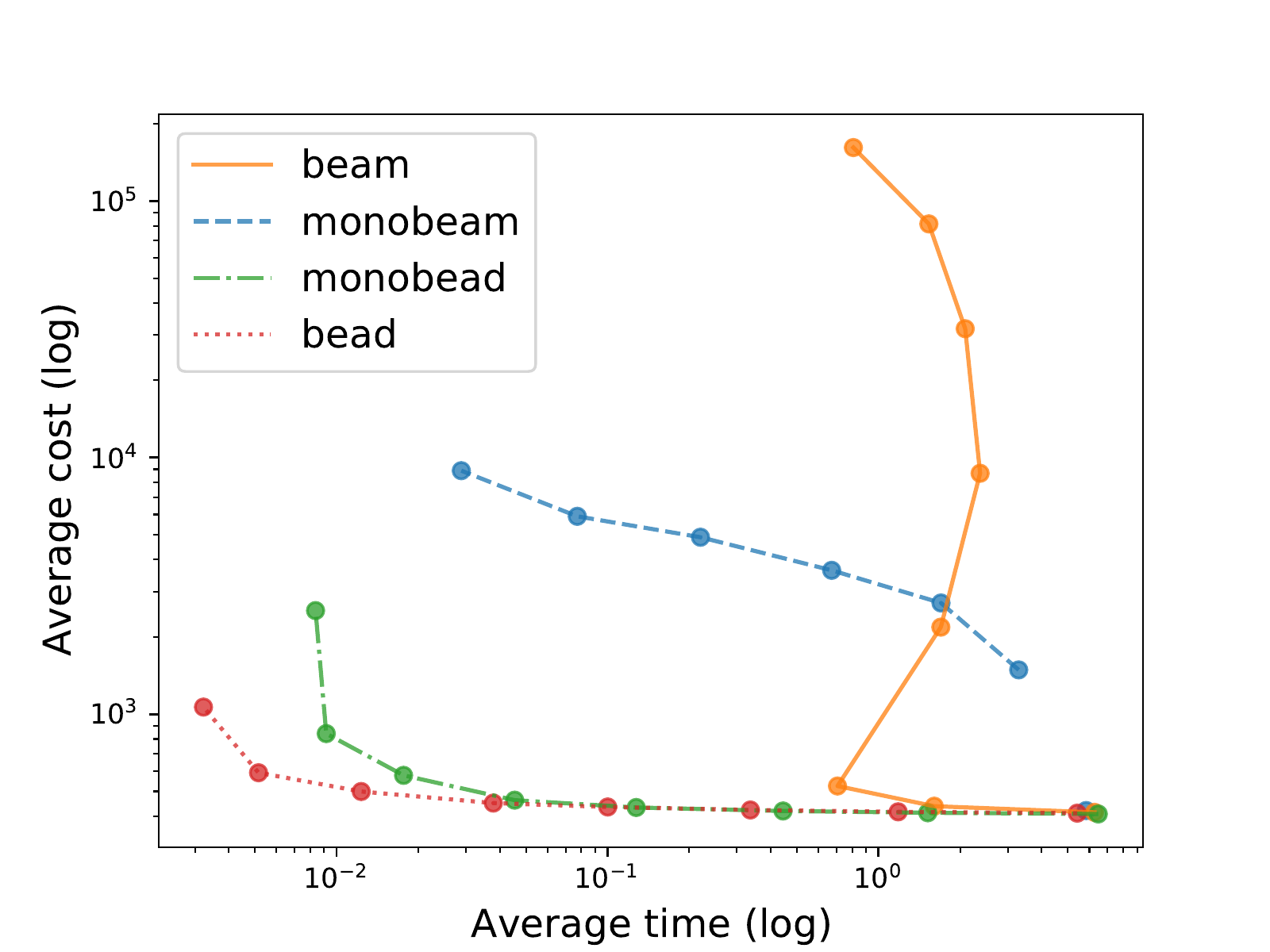}
\vspace{\captionadjust}
\caption{\small{15-puzzle (heavy cost)}}
\label{fig:tile-average-heavy}
\end{subfigure}
\hspace{\plotadjust{}}
\begin{subfigure}{\plotwidth{}}
\incfig{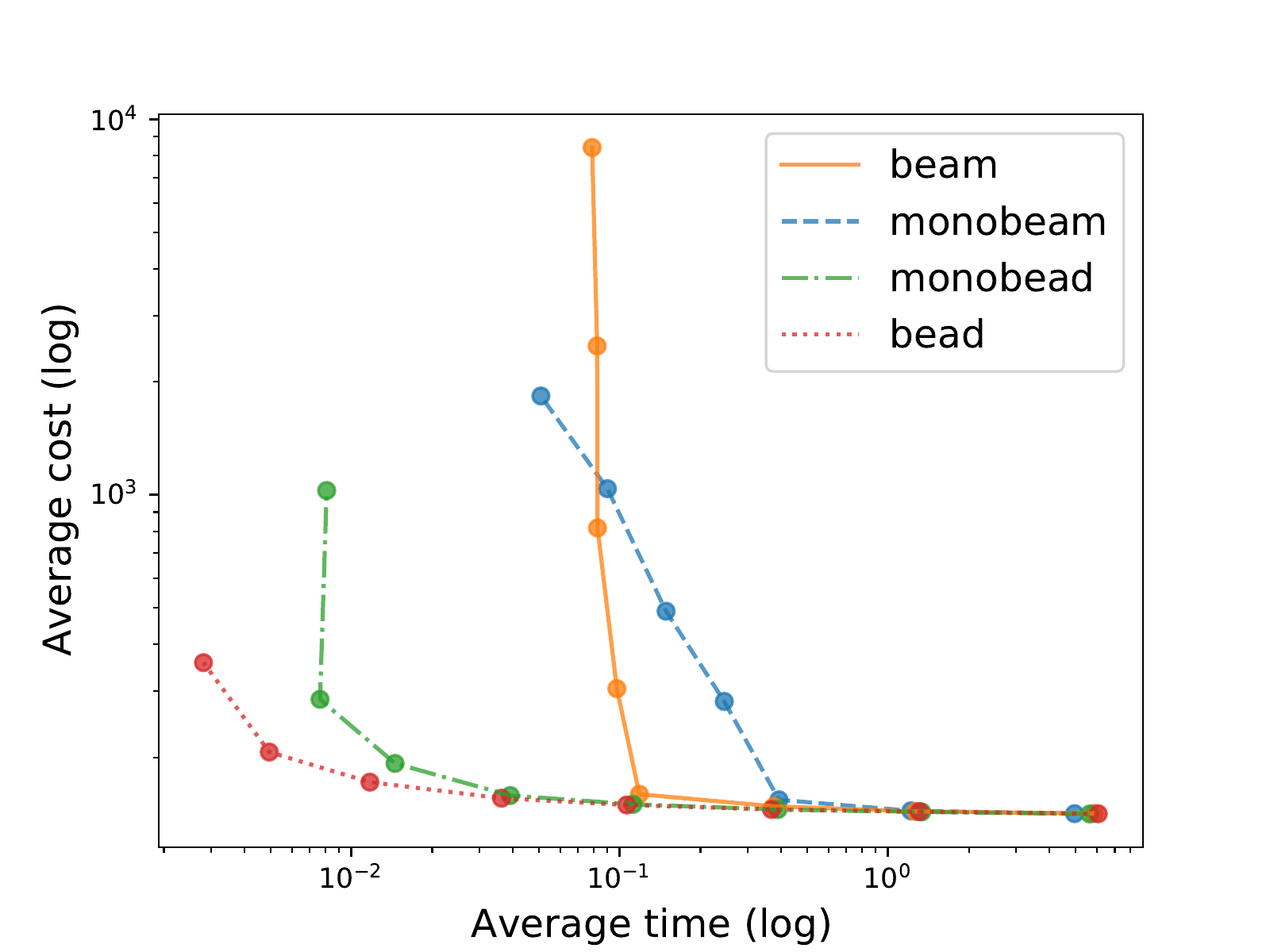}
\vspace{\captionadjust}
\caption{\small{15-puzzle (sqrt cost)}}
\label{fig:tile-average-sqrt}
\end{subfigure}

\begin{subfigure}{\plotwidth{}}
\incfig{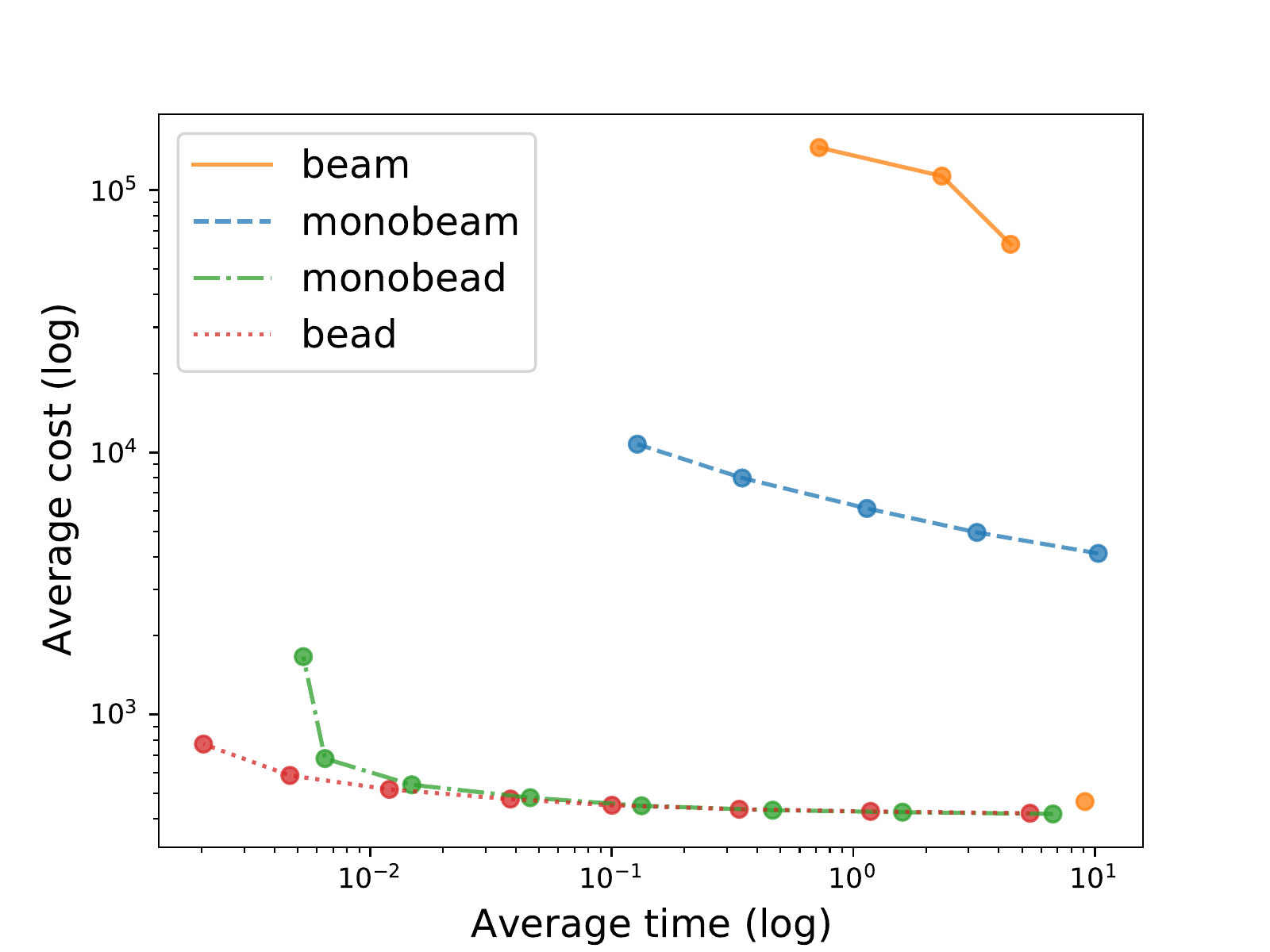}
\vspace{\captionadjust}
\caption{15-puzzle (reverse cost)}
\label{fig:tile-average-reverse}
\end{subfigure}
\hspace{\plotadjust{}}
\begin{subfigure}{\plotwidth{}}
\incfig{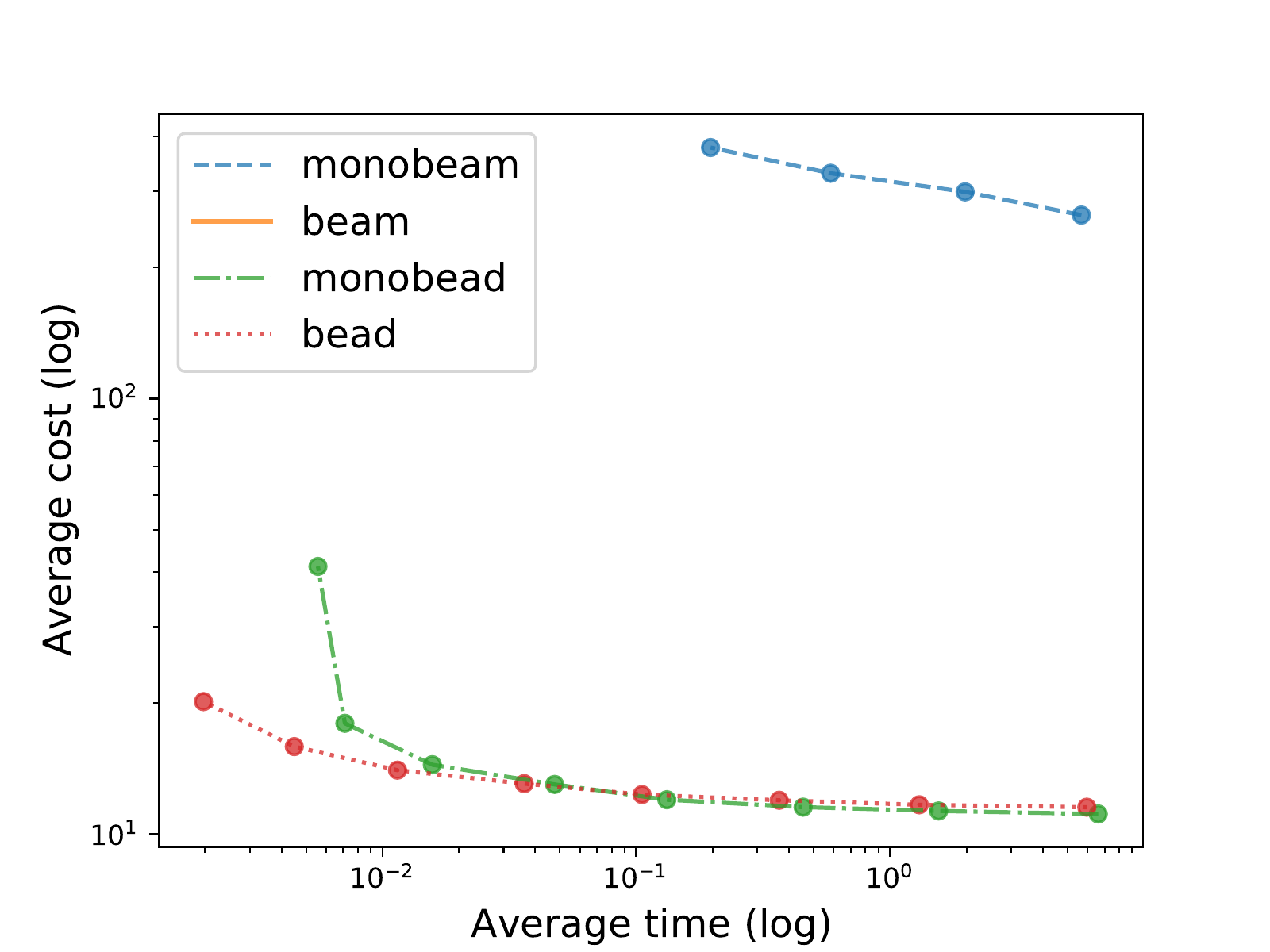}
\vspace{\captionadjust}
\caption{15-puzzle (inverse cost)}
\label{fig:tile-average-inverse}
\end{subfigure}
\hspace{\plotadjust{}}
\begin{subfigure}{\plotwidth{}}
\incfig{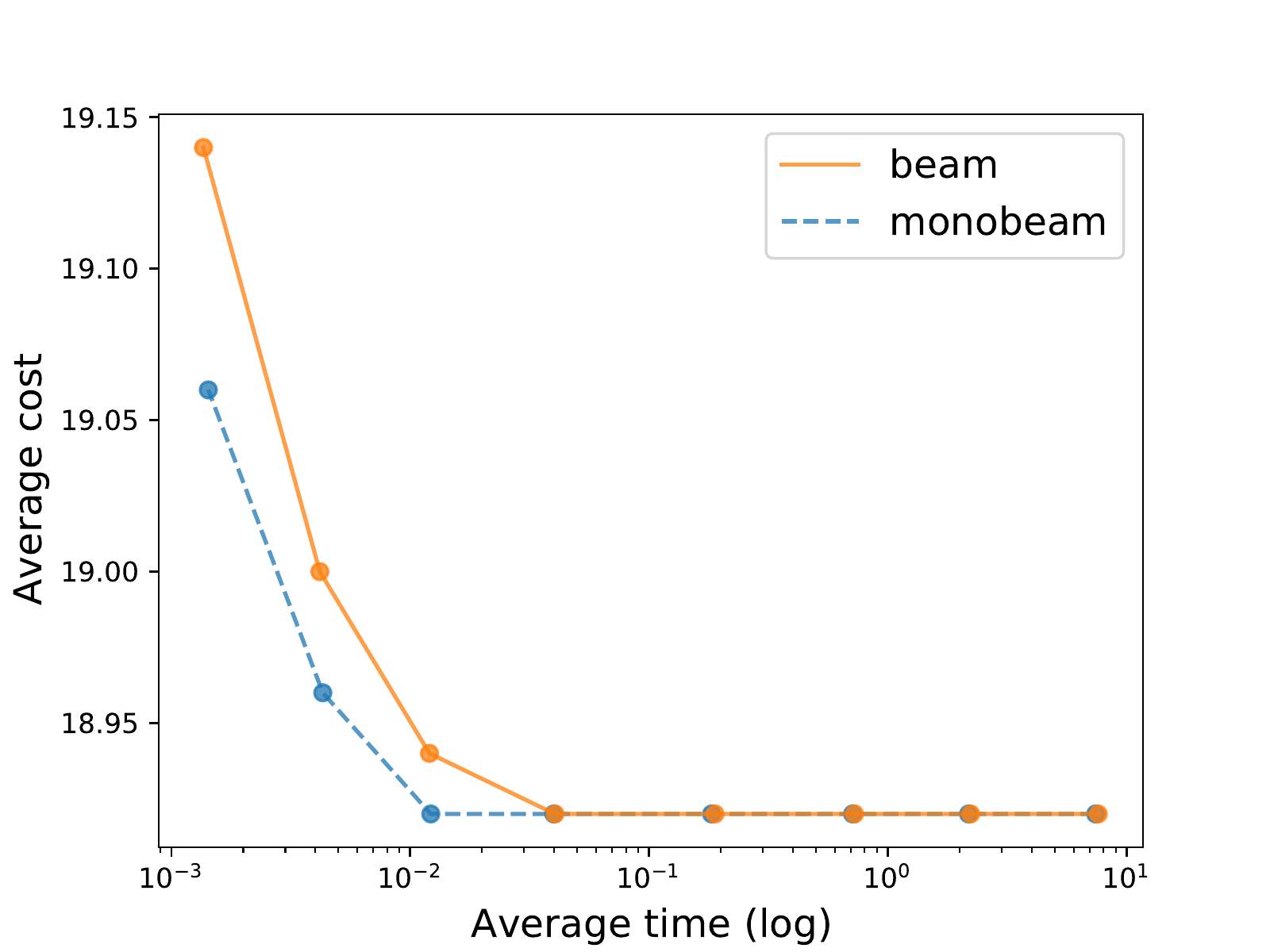}
\vspace{\captionadjust}
\caption{20 pancake problem (unit cost)}
\label{fig:20pancake-average-unit}
\end{subfigure}

\begin{subfigure}{\plotwidth{}}
\incfig{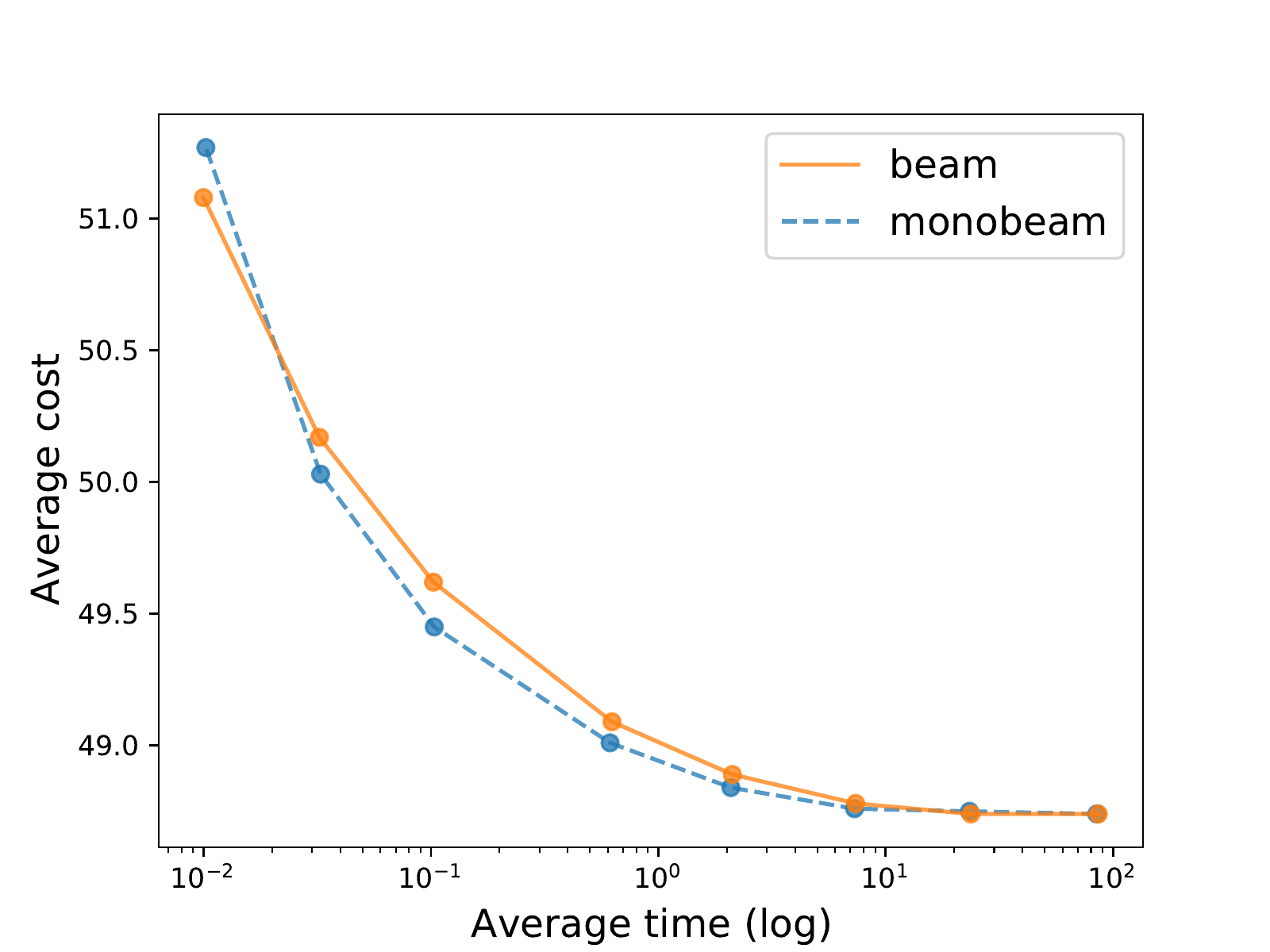}
\vspace{\captionadjust}
\caption{50 pancake problem (unit cost)}
\label{fig:50pancake-average-unit}
\end{subfigure}
\hspace{\plotadjust{}}
\begin{subfigure}{\plotwidth{}}
\incfig{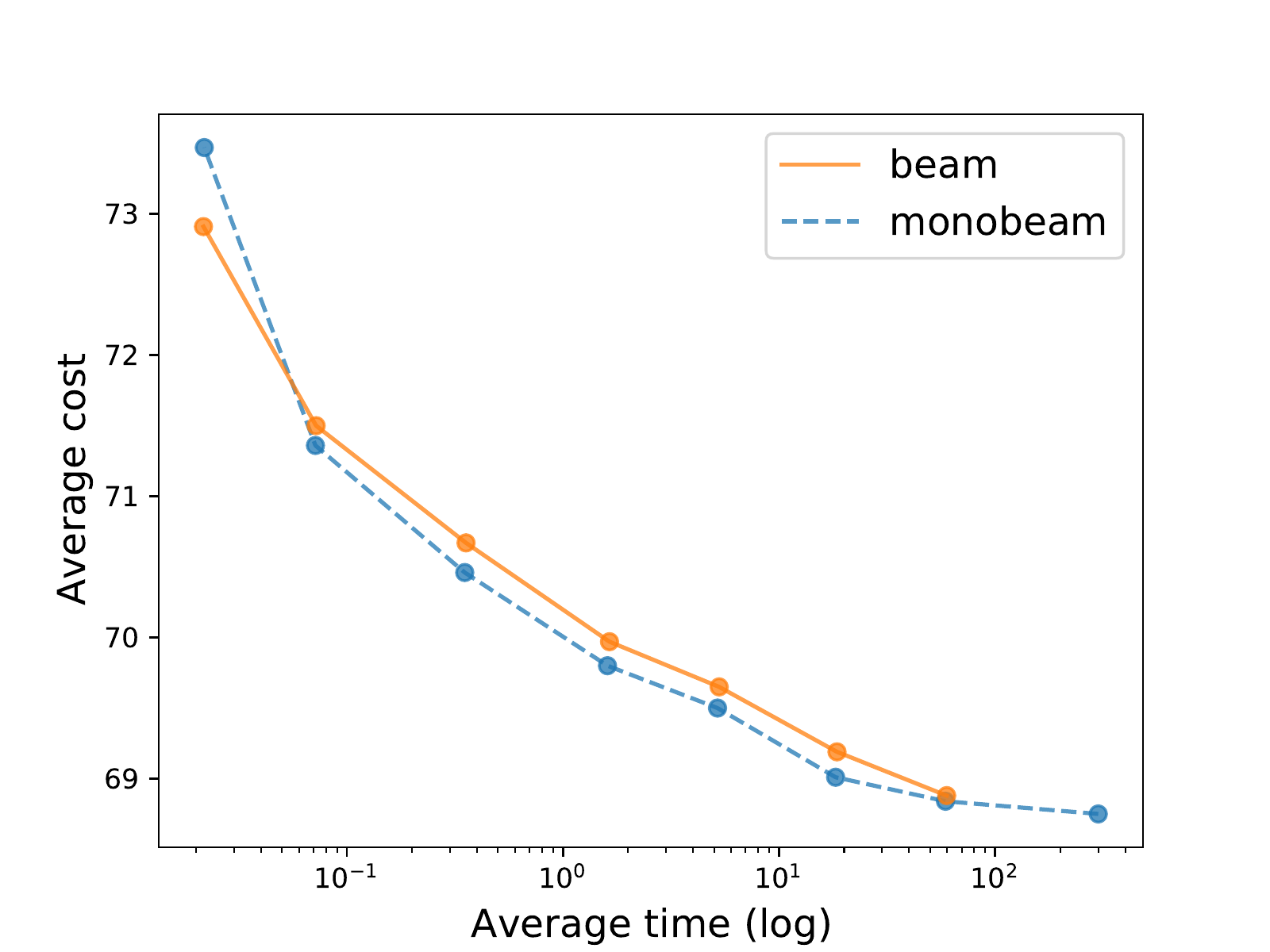}
\vspace{\captionadjust}
\caption{70 pancake problem (unit cost)}
\label{fig:70pancake-average-unit}
\end{subfigure}
\hspace{\plotadjust{}}
\begin{subfigure}{\plotwidth{}}
\incfig{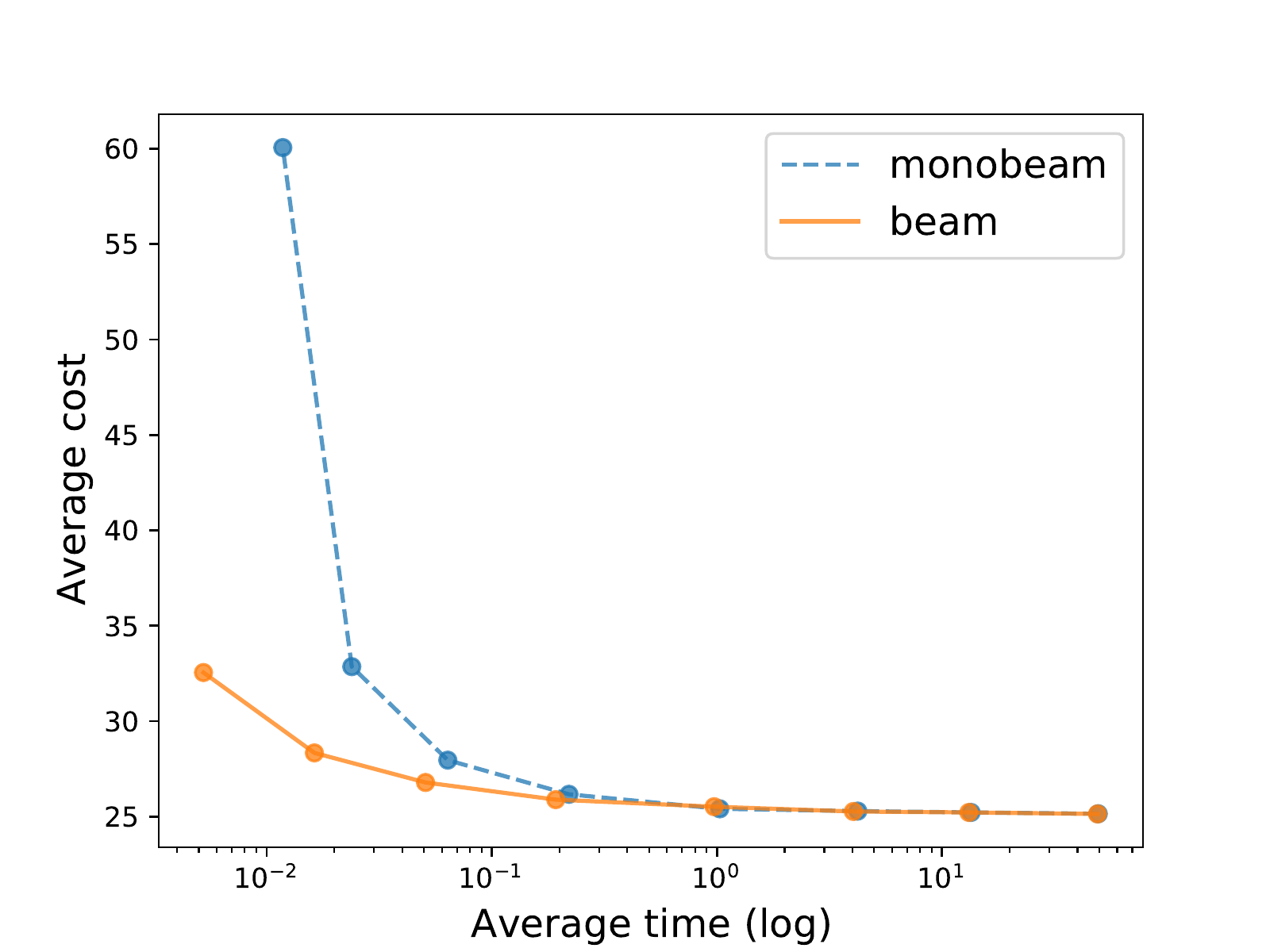}
\vspace{\captionadjust}
\caption{20 blocks world (unit cost)}
\label{fig:20bw-average-unit}
\end{subfigure}

\begin{subfigure}{\plotwidth{}}
\incfig{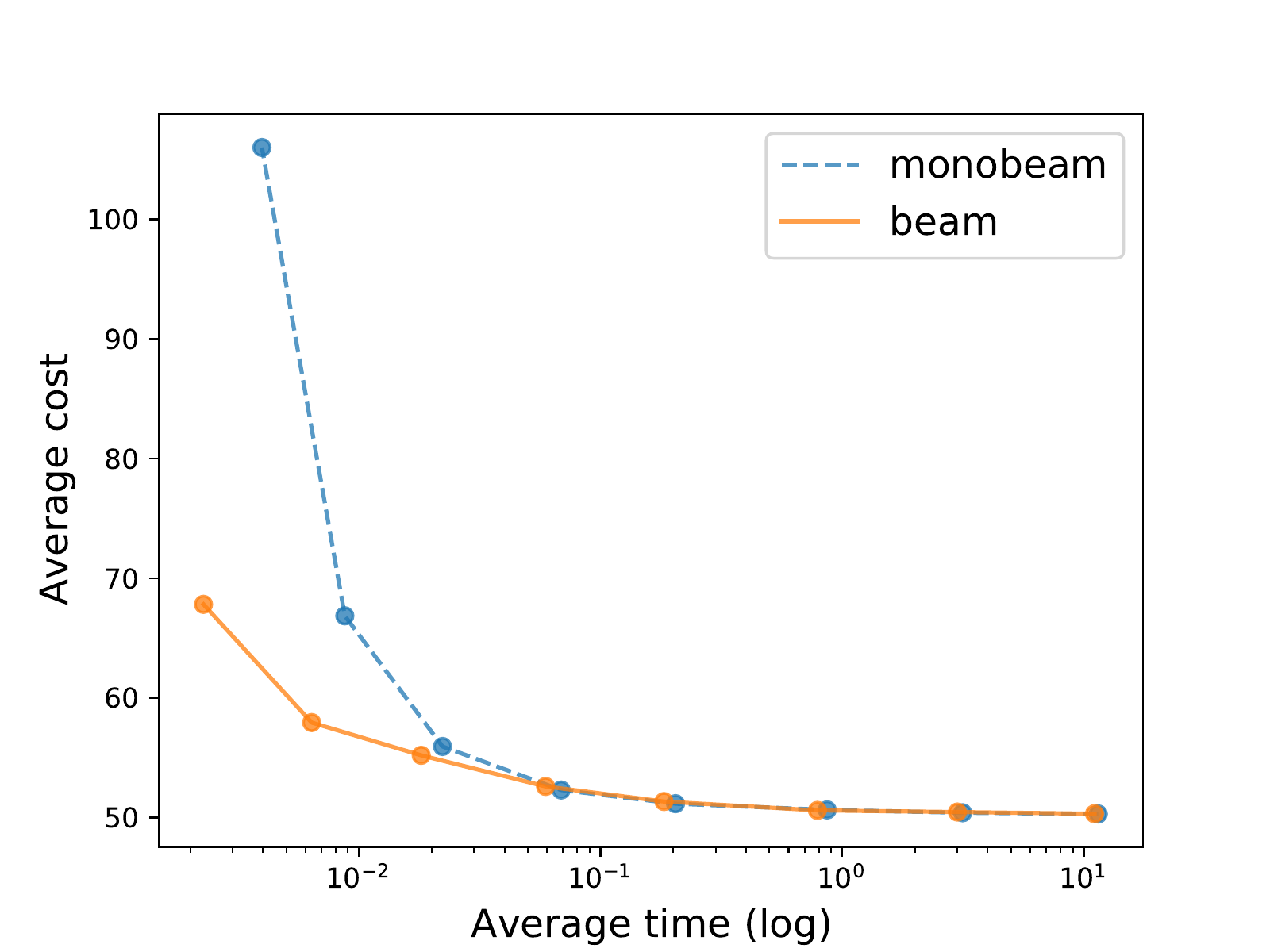}
\vspace{\captionadjust}
\caption{20 deep blocks world (unit cost)}
\label{fig:20bwdp-average-unit}
\end{subfigure}
\hspace{\plotadjust{}}
\begin{subfigure}{\plotwidth{}}
\incfig{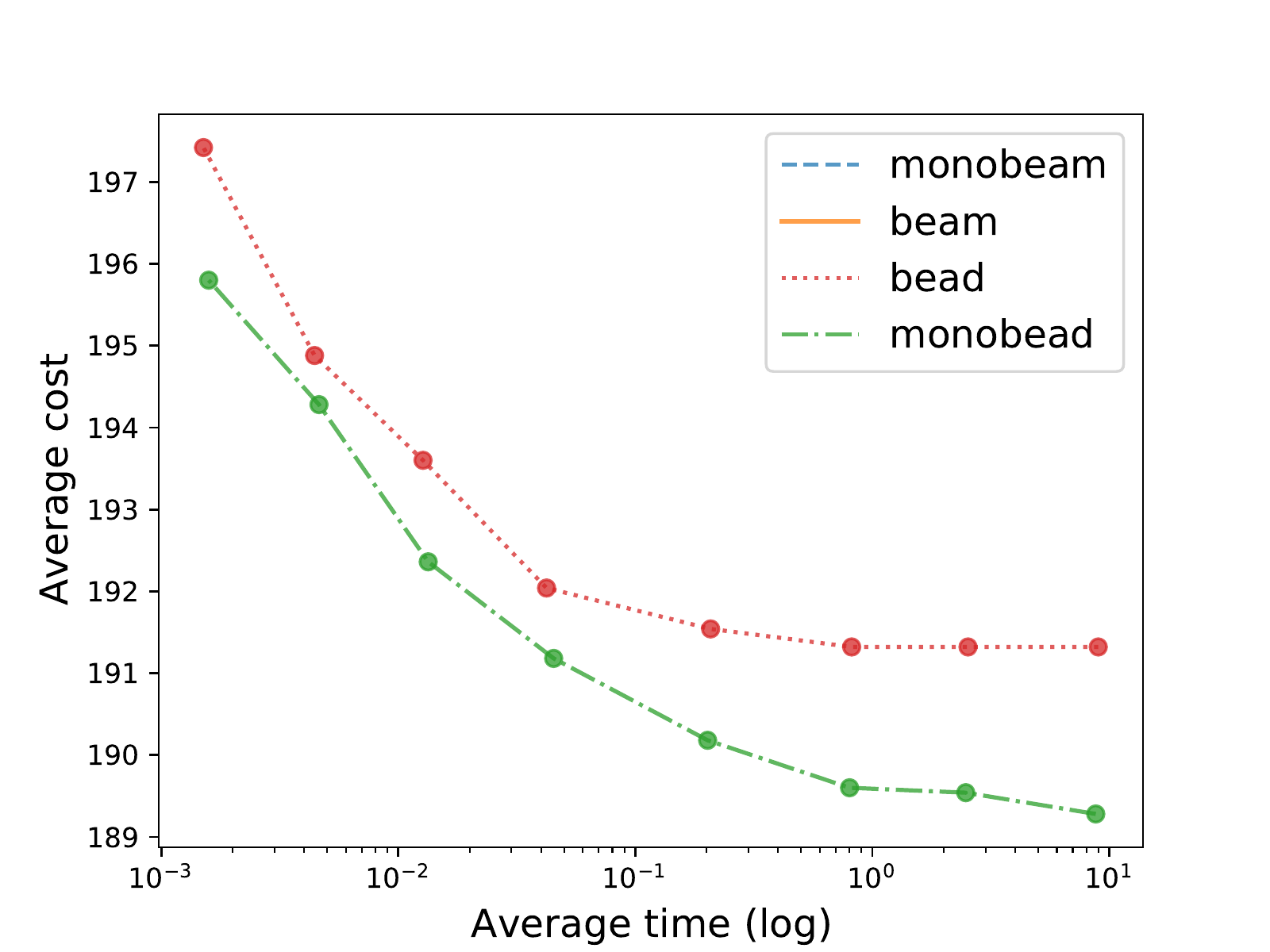}
\vspace{\captionadjust}
\caption{20 pancake problem (heavy cost)}
\label{fig:20pancake-average-heavy}
\end{subfigure}
\hspace{\plotadjust{}}
\begin{subfigure}{\plotwidth{}}
\incfig{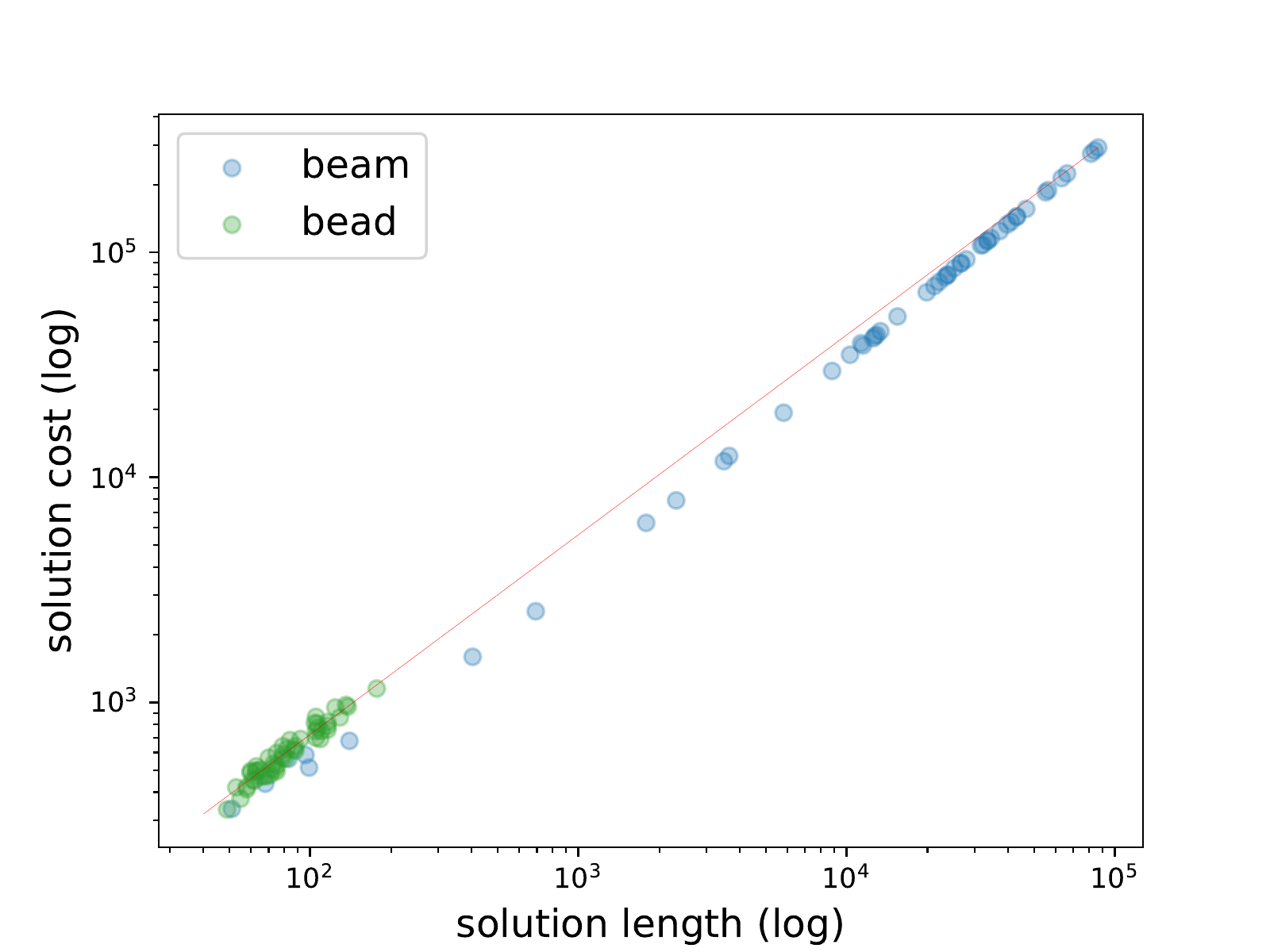}
\vspace{\captionadjust}
\caption{Sol length vs cost (heavy 15-puzzle)}
\label{fig:tile-length-cost}
\end{subfigure}

\caption{Panels a-k: time versus cost as beam width is varied (each point corresponds to a beam width.)  Panel l: solution length versus cost for heavy tiles.}
\label{fig:plots}
\end{figure*}

Next, we study the trade-off between solution cost and solving time in greater detail.
Figure \ref{fig:plots} shows how each algorithm's time / cost trade-off behaves as the beam width is varied.  (Several plots also include bead and monobead, two algorithms that we will discuss below.)  The x axis represents CPU time (log scaled) and the y axis represents the returned solution's cost (often log scaled).  A point is drawn for an algorithm for a specific beam width only if it solved all instances when run with that width. Otherwise, there is simply a gap left in the line for that beam width.
We see in the unit cost plot (Figure \ref{fig:tile-average-unit}) that at the lowest beam widths, monobeam generally finds more expensive solutions and takes more time to do so. But, as the beam width increases, it converges with regular beam search.

The heavy tiles results shown in Figure \ref{fig:tile-average-heavy} demonstrate a different kind of behavior. Beam search is poorly behaved --- at lower beam widths it takes a  long time to return poor quality solutions --- while monobeam solves problems quickly and exhibits a smooth well-behaved improvement with increasing beam width.  At large beam widths (10000 onward), beam finally improves enough to outperform monobeam and at beam width 30000 monobeam fails to solve two instances (due to the increased memory requirements and not finding a solution quickly enough).  Figure \ref{fig:tile-average-sqrt} demonstrates a less dramatic instance of similar behavior, where beam search returns poorer quality solutions for the first two beam widths, and then performs better than or equal with monobeam from there onward.

In reverse cost (Figure \ref{fig:tile-average-reverse}) we see that beam search is unable to reliably solve problems at many beam widths (although eventually does well at the highest beam width of 100000).  Monobeam is again much better behaved, solving problems faster and with lower cost for a wider range of beam widths (it fails only at widths 30, 30000, and 100000). In inverse cost (Figure \ref{fig:tile-average-inverse}), beam search was unable to solve problems reliably at any beam width, while monobeam does well (failing to solve some instances again only at widths 30, 30000, and 100000).  Overall, monobeam is much better behaved on the 15-puzzle than beam.

\subsubsection{Pancake Problem}

We ran with 50 random instances of problems with stacks of 20, 50, and 70 pancakes. The gap heuristic \cite{helmert:2010} was used by all algorithms. The results for the 20-pancake problems, shown in Figure \ref{fig:20pancake-average-unit}, shows monobeam performing on average better than beam search at the lower beam widths, though the differences are small. Likewise, for 50-pancake and 70-pancake problems in Figures \ref{fig:50pancake-average-unit} and \ref{fig:70pancake-average-unit}, we see beam search and monobeam performing roughly the same with minor variations in solution quality.  Overall, it appears that monotonicity might actually benefit solution cost here instead of imposing much of a penalty.

\subsubsection{Blocks World}

We tested on 100 random instances of blocks world, with two different action models: one in which blocks are directly moved to a stack as an action (`blocks world') and one in which picking up and putting down blocks each use an action, so therefore the branching factor is smaller and plans are longer (`deep blocks world').  In both the regular and deep variants (Figures \ref{fig:20bw-average-unit} and \ref{fig:20bwdp-average-unit}), we see a significant trade-off for ensuring monotonicity at lower beam widths (widths 30, 100, and 300) and both algorithms perform comparably at the higher beam widths.

\subsubsection{Summary}

The price of monotonicity appears to be noticeable but modest.  Beam search performs very poorly on non-unit cost domains, with monobeam often behaving much better. However, both algorithms are inconsistent at solving the nefarious reverse and inverse cost 15-puzzles.

\section{Beam Search with Costs}

We saw in the previous section that beam search behaves poorly in domains with non-unit costs.  For best-first search, searching using distance-to-go $d$ (known as speedy search) instead of cost-to-go $h$ (known as greedy best-first search or GBFS) is well-known to yield faster search in non-unit domains \cite{thayer:ude-icaps}.  Originally this was thought to be due to $d$ being a proxy for remaining search effort \cite{ruml:bfu, burns:hsw}.  However, later work suggested that $d$ results in smaller local minima for a best-first search \cite{wilt:svg}.  It is not clear that the concept of a local minimum or crater \cite{heusner:usb} carries over from best-first search to beam search.  However, the results above clearly show poor performance for beam search in non-unit problems.

It is natural to ask whether guiding beam search using $d$ might help.  In this section, we explore this idea.  We develop variants of the algorithms discussed above that prefer nodes with low $d$. First, we introduce {\bf bead} search, which is beam search using a purely distance-based measurement $l(n) = depth(n) + d(n)$ (estimated length of solution) to select nodes for the beam. Bead search uses the depth of the node and not the non-unit-cost $g$-value because the distance-to-go estimate would be overwhelmed by $g$ in domains where costs are much larger than 1. It uses the domain's actual $f$ and $g$ values only for tie-breaking, comparing solutions against an incumbent, and in duplicate elimination for determining if a duplicate node is better than the previously seen version.

We also introduce {\bf monobead} search, which is monotonic beam search according to the algorithm given in Figures \ref{alg:monobeam} and \ref{alg:dedup}, except using $l(n)$ to select nodes for the beam. As with bead search, monobead uses the domain's $f$ and $g$ values for tie breaking and duplicate elimination. It does not use incumbent-based pruning, because Theorem \ref{theo:pruning} only applies when the search is ordered on the same value function used for pruning.

\subsection{Empirical Analysis}

\subsubsection{Sliding Tile Puzzle}
We use the unweighted Manhattan distance for the distance-to-go estimate $d$ used by bead and monobead.

In Figure \ref{fig:tile-average-heavy}, we see that in the heavy tiles domain, the d-based algorithms, bead and monobead, find solutions more quickly, more consistently, and with lower solution cost than the corresponding $f$-based algorithms, beam, and monobeam. Note that there is no gap in the line for monobead, meaning that it solves all instances at all beam widths, unlike monobeam. And Figure \ref{fig:tile-average-inverse} shows an even more pronounced case of this, with bead and monobead performing very reasonably whereas beam search never solved more than 80\% of the problems at any beam width and monobeam failed to solve around 20\% of the instances at beam width 30 and failed to solve 20-30\% of the instances at the two highest beam widths (30000 and 100000).  In all the non-unit 15-puzzle cost models, the $d$-based variants are clearly the best performing.

Again, we observe that at lower beam widths, monobead generally finds less desirable solutions than bead and takes more time to do so. But, as the beam width increases, solution quality converges to about the same than bead (and slightly better in inverse). Bead delivers best solutions under 0.01 seconds but when larger beam widths are used, the solution delivered by both bead and monobead converges and they both take the same time ---see Figures~\ref{fig:tile-average-unit}--\ref{fig:tile-average-inverse}.

\subsubsection{Pancake Problem}

The algorithms were tested on 50 random instances of the heavy pancake problem \cite{hatem:2014}, in which each pancake is given an ID number from 1 through $N$ (the number of pancakes), and the cost of a flip is the ID of the pancake above the spatula. An adapted version of the gap heuristic was used for cost-to-go estimates, and the standard gap heuristic was used as a distance-to-go estimate $d$ for bead and monobead.

Figure~\ref{fig:20pancake-average-heavy} shows the performance of the algorithms in this domain.  The non-$d$ variants were not able to solve all instances for any of the beam widths and so there are no points displayed for them in the figure. Monobeam is able to solve some instances at beam widths 100 through 30000, but never above 25\% of all instances tested. However, the $d$-based variants are able to consistently solve all instances and improve their solution quality as beam width increases.  The monotonic algorithm (monobead) performs better than the non-monotonic (bead) algorithm, though the differences in cost are relatively small.

\subsubsection{Summary}

As would be expected, searching on distance-to-go in non-unit domains leads us to solutions faster than searching on cost-to-go.  This effect is so powerful that it allows us to find solutions where we otherwise could not. More surprisingly, we see that beam searches using distance-to-go generally find solutions with much lower cost than their cost-to-go counterparts.  Indeed, when solving heavy 15-puzzles with a beam width of 100, beam search typically found solutions 25533 steps long, resulting in an average cost of 85843, while bead found solutions of length 86, resulting in an average cost of only 622.

We speculate that this is due to a correlation between solution length and cost.  Figure \ref{fig:tile-length-cost} shows the costs and lengths of the solutions found by beam and bead.  The red line indicates what the cost would be if a solution of a particular length used actions of average cost (8 for heavy tiles).  We see that bead finds much shorter solutions, although they tend to be more expensive per action than those found by beam. One way to understand this is that, because $d$-based beam searches ignores action costs (except for duplicates, pruning, and tie-breaking) and instead focus on solution length, their solutions will tend to be short.  If actions along short solutions aren't unusually expensive, then the solutions found by $d$-based search will have costs roughly proportional to the shortest solution path times the average action cost.  This can be much cheaper than the arbitrarily long solutions that are found by $f$-guided beam searches.

\section{Discussion}

Although this work provides a better-behaved beam search whose solution cost is monotonic in beam width, clearly much work remains before we fully understand the behavior of beam search.  Our analysis of $d$-based beam search suggests that, in the domains we tested, it can find short solution that nonetheless use actions of only average cost, but this phenomenon deserves comprehensive scrutiny.  It remains important to fully understand why $f$-guided beam search fails to find the shorter and much cheaper solutions that bead does.

The monobeam algorithm has the limitation that when nodes can have no children (for example, dead ends) or when pruning from duplicate elimination or a candidate solution causes no children to be kept from a node, there may be times at which no node is available to select for the current beam slot. If all slots below the current one are also empty, those slots will stay empty for the remainder of the search. For example, if slot 1 in our beam empties at one level of the search, we will not be able to expand a node from slot 1 to fill it ever again. A method to address this problem could improve the performance of the algorithm, but it is not obvious how to retain a guarantee of monotonicity, as any node that fills in the gap may well act as a cuckoo node.

Our benchmarks include several classic combinatorial search domains, but puzzle-like domains are overrepresented and it would be beneficial to extend our study to domains of different character.

The other major family of unboundedly-suboptimal heuristic search algorithms is based on best-first search.  There has been work on understanding GBFS \cite{heusner:usb} and making it more robust \cite{valenzano:cke, xie:uil}.
As mentioned above, distance-to-go has been found an effective accelerator in the best-first setting \cite{thayer:ude-icaps} as it appears to generate smaller local minima \cite{wilt:svg}.  
It is an important area of future work to understand the analog of local minima for beam search and explain the behavior of $h$ versus $d$ in this setting.

\section{Conclusions}

We have advanced the study of beam search in two ways.  First, we introduced a simple monotonic variant of beam search, monobeam, and studied its interaction with duplicate detection.  Monobeam tends to find slightly worse solutions for small widths in unit-cost problems, but is more robust and easier to use.  Second, we proposed using distance-to-go to guide beam search, resulting in new variants that perform much better in non-unit-cost domains.  Given the importance of scalability in many applications, we hope this work widens the applicability of heuristic search to problems of practical importance.

\bigskip

\section*{Acknowledgments}

We are grateful for support from the NSF-BSF program (via NSF grant 2008594) and Earlham College (via the Lemann Student/Faculty Collaborative Research Fund) and to Liam Peachey for work on the initial blocks world implementation.

\bibliography{master}

\end{document}


\maketitle

\fix{need to add template and check formatting for supplemental-mat.tex}

\section{Beam Search}

Pseudocode for our implementation of beam is shown in Algorithm~\ref{alg:beam}.  Beam search takes as parameters the start state and beam width, and searches a state space level by level, as in breadth-first search. Duplicate detection is done by comparing a newly generated node (line~\ref{lin:dup}) to  the contents of previous beams (line~\ref{lin:addtoclosed}).
If more nodes are generated for the next level than allowed by the beam width, only those with the lowest f-values are retained (line~\ref{lin:beamsel}).  Ties are broken in favor of nodes with lower h-values.  The algorithm stops after completing the level at which a solution is found (line~\ref{lin:beamloop}), returning the lowest cost solution from that level if there are multiple (line~\ref{lin:beamsol}). Everything about our implementation is standard except the last point: some implementations stop as soon as a solution is found.

\fix{(need to cite: David Furcy's thesis, Figure 39, page 103).}

\begin{algorithm}[h!]
\Begin{
solutionCost $\leftarrow \infty$\;
beam[1] $\leftarrow$ start\;
closed $\leftarrow \emptyset$\;
\While {at least one slot in the beam has a node and no solution found}{\label{lin:beamloop}
candidates $\leftarrow \emptyset$\;
\For{each beam slot c from 1 to width}{
\If {beam[c] is a node}{
\For{each child of beam[c]}{
\If {child is a goal}{
\If{$f$(child) $<$ solutionCost}{\label{lin:beamsol}
store as solution\;
solutionCost $\leftarrow f$(child)\;
}
}
\Else{
\If{child not in closed or $f$(child) $< f$(closed entry)\label{lin:dup}}{
add child to candidates\;
}}
}
}
}
beam $\leftarrow [ ]$\;
\For{each beam slot c from 1 to width}{
\If {candidates is nonempty}{
beam[c] $\leftarrow$ remove $\min f$-value node from candidates\;\label{lin:beamsel}
add beam[c] to closed\label{lin:addtoclosed}\;
}
}
}
return solution\;
}
\caption{Beam(start,width)}\label{alg:beam}
\end{algorithm}